\documentclass[10pt,twocolumn,letterpaper]{article}

\usepackage{cvpr}
\usepackage{times}
\usepackage{epsfig}
\usepackage{graphicx}
\usepackage{subfigure}
\usepackage{amsmath}
\usepackage{amssymb}
\usepackage[OT1]{fontenc}
\usepackage{indentfirst}
\usepackage{bm}
\usepackage{xspace}
\usepackage{float}
\usepackage{lipsum}
\usepackage{caption}
\usepackage{titling}

\usepackage[pagebackref=true,breaklinks=true,letterpaper=true,colorlinks,bookmarks=false]{hyperref}

\cvprfinalcopy 

\begin{document}

\title{Less Is More: Picking Informative Frames for Video Captioning}

\author{Yangyu Chen$^{1}$, Shuhui Wang$^{2}$, Weigang Zhang$^{3}$ and Qingming Huang$^{1}$\\
$^{1}$University of Chinese Academy of Science, Beijing 100049, China\\
$^{2}$Institute of Computing Technology, Chinese Academy of Sciences, Beijing 100190, China\\
$^{3}$Harbin Institute of Technology, Weihai 264209, China\\
{\tt\small \{yangyu.chen, shuhui.wang, weigang.zhang, qingming.huang\}@vipl.ict.ac.cn}
}
\date{}

\setlength{\droptitle}{-2.5em}
\maketitle

\begin{abstract}
  In video captioning task, the best practice has been achieved by attention-based models which associate salient visual components with sentences in the video. However, existing study follows a common procedure which includes a frame-level appearance modeling and motion modeling on equal interval frame sampling, which may bring about redundant visual information, sensitivity to content noise and unnecessary computation cost.
  
  We propose a plug-and-play \textsf{PickNet} to perform informative frame picking in video captioning. Based on a standard Encoder-Decoder framework, we develop a reinforcement-learning-based procedure to train the network sequentially, where the reward of each frame picking action is designed by maximizing visual diversity and minimizing textual discrepancy.  If the candidate is rewarded, it will be selected and the corresponding latent representation of Encoder-Decoder will be updated for future trials. This procedure goes on until the end of the video sequence. Consequently, a compact frame subset can be selected to represent the visual information and perform video captioning without performance degradation. Experiment results shows that our model can use $6 \scriptsize{\sim}8$ frames to achieve competitive performance across popular benchmarks.
\end{abstract}
\vspace{-1ex}

\section{Introduction}
\vspace{-1ex}
Human are born with the ability to identify useful information and filter redundant information.
In biology, this mechanism is called sensory gating~\cite{Cromwell2008}, which describes neurological processes of filtering out unnecessary stimuli in the brain from all possible environmental stimuli, thus prevents an overload of redundant information in the higher cortical centers of the brain. This cognitive mechanism is essentially consistent with a huge body of researches in computer vision.

As one of the strong evidences practicing on visual sensory gating, attention is introduced to identify the salient visual regions with high objectness and meaningful visual patterns of an image~\cite{Mnih2014b, Zheng2017a}. The attention has also been established on videos that contains consecutive image frames. Existing study follows a common procedure which includes a frame-level appearance modeling and motion modeling on equal interval frame sampling, say, every 3 frames or 5 frames. Visual features and motion features are extracted on the selected frame subset one by one, and they are all fed into the learning stage. Similar to image, the video attention is recognized as a spatial-temporal saliency that identifies both salient objects and their motion trajectories~\cite{Shen2017}. The video attention is also recognized as the word-frame association learned by sparse coding~\cite{Yao2015a} or gaze-guided attention learning~\cite{Yu2017}, which is a de-facto frame weighting mechanism.
The visual attention mechanism also benefits many downstream tasks such visual captioning and visual question answering for image and video~\cite{Yang2015, Lu2016a, You2016, Hori2017}.

\begin{figure}[tb]
  \begin{center}
    \subfigure[Equally sampled 30 frames from a video]{ \includegraphics[width=\linewidth]{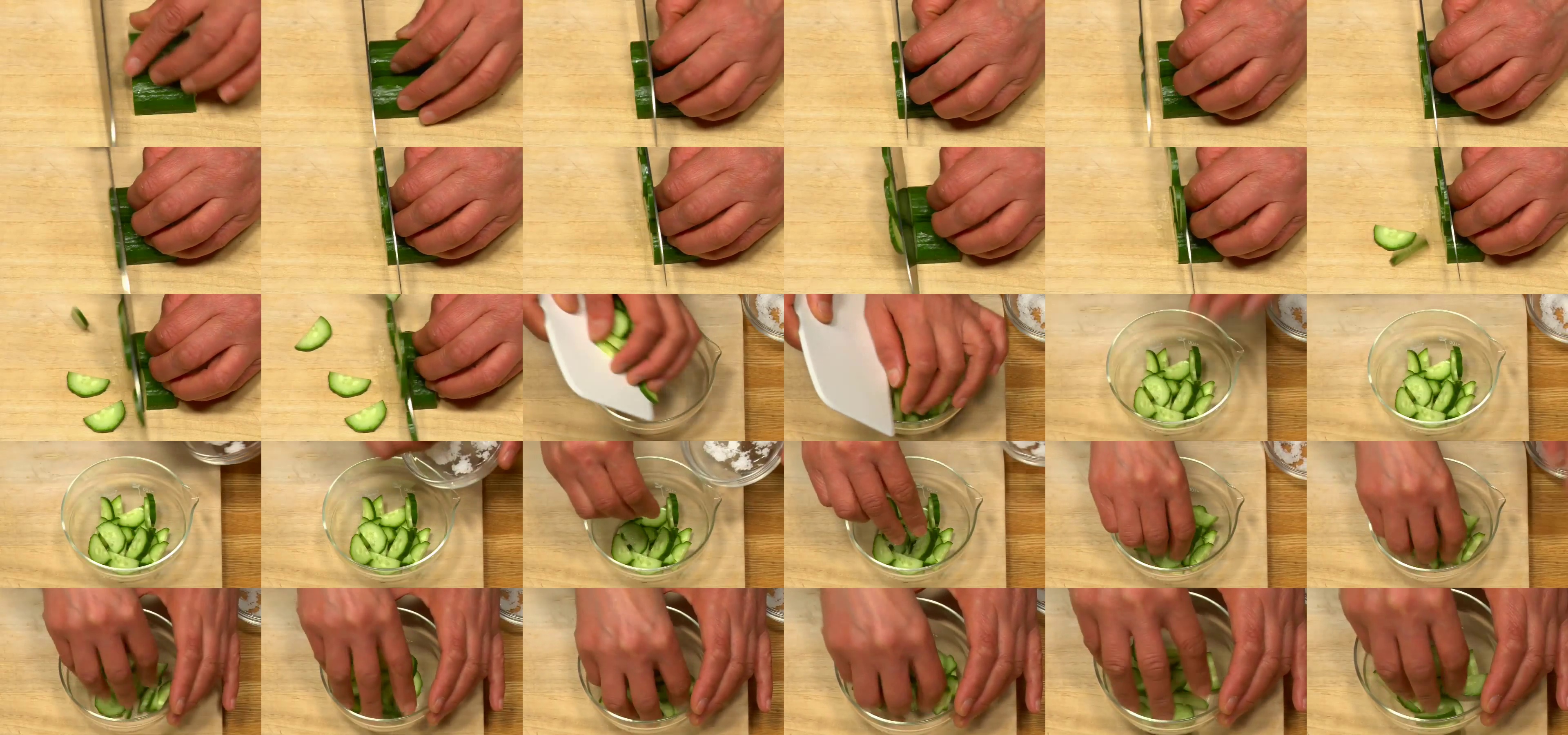} } \\
    \vspace{-0.6em}
    \subfigure[Informative frames]{ \includegraphics[width=\linewidth]{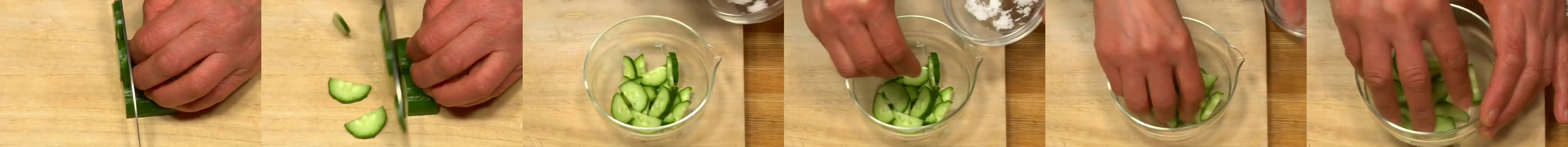} }
  \end{center}
\vspace{-2em}
\caption{An illustration of the temporal redundancy in video. Video may contains many redundant information. Using only a portion of frames can clearly convey information.}
\label{fig:redundant}
\end{figure}

Despite the success on bridging vision and language achieved by existing attention-based methods, there still exists critical issues to be addressed as follows.
\begin{itemize}
  \setlength\itemsep{0.2em}
    \item {\bf Frame selection perspective.} As shown in Figure~\ref{fig:redundant}(a), there are many frames with duplicated and redundant visual appearance information selected with equal interval frame sampling. This will also involve remarkable computation expenditures and less performance gain as the information from the input is not appropriately sampled. For example, it takes millions of floating point calculation to extract a frame-level visual feature for a moderate-sized CNN model. Moreover, there is no guarantee that all the frames selected by equal interval sampling contain meaningful information, so it tends to be more sensitive to content noise such as motion blur, occlusion and object zoom-out.
    \item {\bf Downstream video captioning task perspective.} Previous attention-based models mostly identify the spatial layout of visual saliency, but the temporal redundancy existing in neighboring frames remains unsolved as all the frames are taken into consideration. This may lead to an unexpected information overload on the visual-linguistic correlation analysis model. For example, the dense-captioning-based strategy~\cite{Krause2016,Johnson2016,Shen2017} can potentially describe images/videos in finer levels of detail by captioning many visual regions within an image/video-clip. With an increasing number of frames, many highly similar visual regions will be generated and the problem will become prohibitive as the search space of sequence-to-sequence association becomes extremely large.
\end{itemize}

We answer the follow question: \textit{Is there a way to use as less number of frames as possible to well approximate the performance using all the frames for video captioning?} We propose \textsf{PickNet} to perform informative frame picking for video captioning. Specifically, the base model for visual-linguistic association in video captioning is a standard Encoder-Decoder framework~\cite{Baraldi2016}. We develop a reinforcement-learning-based procedure to train the network sequentially, where the reward of each frame picking action is designed by considering both visual and textual cues. From visual perspective, we maximize the diversity between current picked frame candidate and the selected frames. From textual perspective, we minimize the discrepancy between ground truth caption and the generated one using current picked candidate. If the candidate is rewarded, it will be selected and the corresponding latent representation of Encoder-Decoder will be updated for future trials. This procedure goes on until the end of the video sequence. Consequently, a compact frame subset can be selected to represent the visual information and perform video captioning without performance degradation.

To the best of our knowledge, this is the first study on frame selection for video captioning. In fact, our framework can go beyond the Encoder-Decoder framework in video captioning task, and serves as a complementary building block for other state-of-the-art solutions. It can also be adapted by other task-specific objectives for video analysis. In summary, the merits of our \textsf{PickNet} include:
\begin{itemize}
\setlength\itemsep{0.2em}
\item {\bf Flexibility.} We design a plug-and-play reinforcement-learning-based \textsf{PickNet} to select informative frames which can pick informative frames for the next learning stage. A compact frame subset can be selected to represent the visual information and perform video captioning without performance degradation.
\item {\bf Efficiency.} The architecture can largely cut down the usage of convolution operations. It makes our method more applicable for real-world video processing.
\item {\bf Effectiveness.} Experiment shows that our model can use a small number of frames to achieve comparable or even better performance compared to state-of-the-art.
\end{itemize}
\vspace{-1ex}

\section{Related Works}
\vspace{-1ex}
\subsection{Visual captioning}
The visual captioning is the task that translating visual contents into natural language. Early to 2002, Kojima \textit{et al.}~\cite{Kojima2002} proposed the first video captioning system for describing human behavior. From then on, a series of image and video captioning studies have been conducted. Early approaches tackle this problem using bottom-up paradigm~\cite{Farhadi2010, Kulkarni2011, Yang2011, Fang2015}, which first generate descriptive words of an image by attribute learning and object recognition, then combine them by language models which fit predicted words to predefined sentence templates. With the development of neural networks and deep learning, modern captioning systems are based on CNN and RNN, with the Encoder-Decoder architecture.

An active branch of captioning is utilizing the attention mechanism to weight the input features. For image captioning, the mechanism is typically in the form of spatial attention. Xu \textit{et al.}~\cite{Xu2015} first introduced an attention based model that automatically learn to fix its gaze on salient objects while generating the corresponding words in the output sequence. For video captioning, the temporal attention is added. Yao \textit{et al.}~\cite{Yao2015a} took into account both the local and global temporal structure of videos to produce descriptions, and their model learned to automatically select the most relevant temporal segments given the text-generating RNN. However, the attention based methods, especially temporal attention, are operated on full observed condition, which is not suitable in some real world applications, such as blind navigation. Our method do not require the global information of videos, which is more effective in these applications.

\subsection{Frame selection}
The main battle of studying how to select informative frames of videos is in the video summarization field. This problem may be formulated as image searching. For example, Song \textit{et al.}~\cite{Song2015} considered images related to the video title that can serve as a proxy for important visual concepts, so they developed a co-archetypal analysis technique that learns canonical visual concepts shared between video and images, and used it to summarize videos.  Other people use sparse learning to deal with this problem. Zhao \textit{et al.}~\cite{Zhao2014} proposed to learn a dictionary from given video using group sparse coding, and the summary video was then generated by combining segments that cannot be sparsely reconstructed using the learned dictionary.

Some video analysis task cooperates with frame selection mechanism. For example, in action detection, Yeung \textit{et al.}~\cite{Yeung2016} designed a policy network to directly predict the temporal bounds of actions, which decreased the need for processing the whole video, and improved the detection performance. However, the prediction made by this method is in the form of normalized global position, which requires the knowledge of the video length, and this makes it unable to deal with real video streams. Different from the above methods, our model select frames based on both semantic and visual information, and do not need to know the global length of videos.
\vspace{-2ex}

\section{Method}
\vspace{-1ex}
Our method can be viewed as the combination of two parts: the Encoder-Decoder based sentence generator and the \textsf{PickNet}.

\subsection{Preliminary}
Like most of video captioning methods, our model is built on the Encoder-Decoder based sentence generator. In this subsection, we briefly introduce this building block.
\begin{figure}[H]
  \centering
  \includegraphics[width=\linewidth]{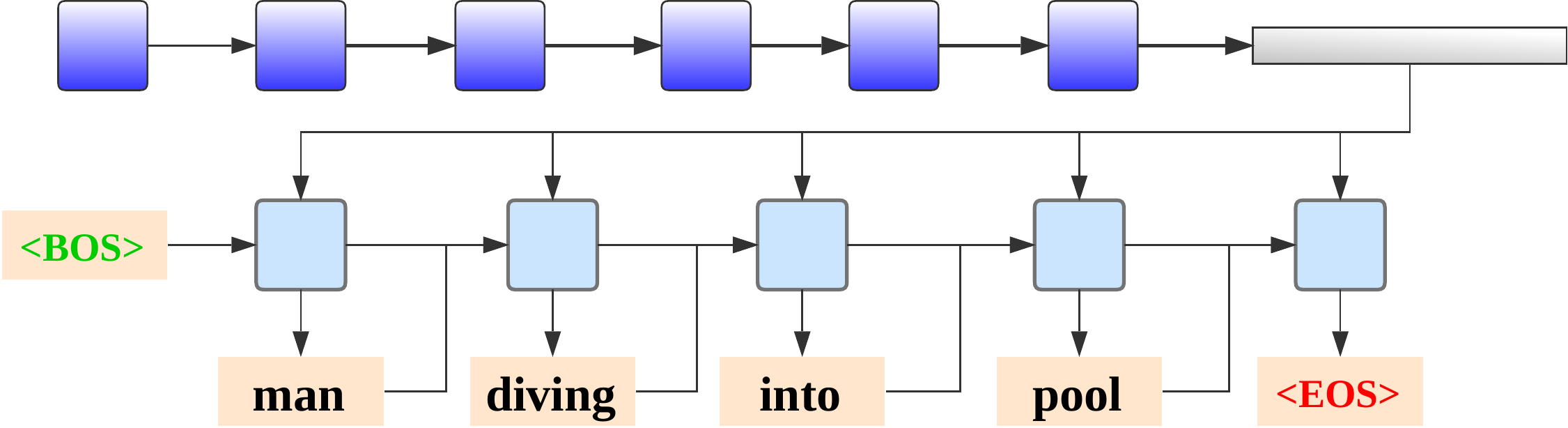}
\vspace{-1em}
\caption{The encode-decode procedure for video captioning.}
\label{fig:encoderdecoder}
\end{figure}

\noindent \textbf{Encoder.} Given an input video, we use a recurrent video encoder which takes a sequence of visual features $(\mathbf{x}_{1}, \mathbf{x}_{2}, \ldots, \mathbf{x}_{n})$ as input and outputs a fixed size vector $\mathbf{v}$ as the representation of this video. The encoder is built on top of a Long Short-Term Memory (LSTM)~\cite{hochreiter1997} unit, which has been widely used for video encoding, since it is known to properly deal with long range temporal dependencies. Different from vanilla recurrent neural network unit, LSTM introduces a memory cell $\mathbf{c}$ which maintains the history of the inputs observed up to a time-step. Specifically, we use the following equations:
\vspace{-0.54em}
\begin{align}
\mathbf{i}_t &= \sigma(W_{ix} \mathbf{x}_t + W_{ih} \mathbf{h}_{t-1} + \mathbf{b}_i) \\
\mathbf{f}_t &= \sigma(W_{fx} \mathbf{x}_t + W_{fh} \mathbf{h}_{t-1} + \mathbf{b}_f) \\
\mathbf{\tilde{c}}_t &= \phi(W_{gx} \mathbf{x}_t + W_{gh} \mathbf{h}_{t-1} + \mathbf{b}_g) \\
\mathbf{c}_t &= \mathbf{f}_t \odot \mathbf{c}_{t-1} + \mathbf{i}_t \odot \mathbf{\tilde{c}}_t \\
\mathbf{o}_t &= \phi(W_{ox} \mathbf{x}_t + W_{oh} \mathbf{h}_{t-1} + \mathbf{b}_o) \\
\mathbf{h}_t &= \mathbf{o}_t \odot \phi(\mathbf{c}_t),
\label{eq:lstm}
\end{align}
where $\odot$ denotes the element-wise Hadamard product, $\sigma$ is the sigmoid function, $\phi$ is the hyperbolic tangent \texttt{tanh}, $W_{*}$ are learned weight matrices and $\mathbf{b}_*$ are learned biases vectors. The hidden state $\mathbf{h}$ and memory cell $\mathbf{c}$ are initialized to zero. And the last hidden state $\mathbf{h}_{T}$ is used as the final encoded video representation $\mathbf{v}$.

\noindent \textbf{Decoder and sentence generation.} Once the representation of the video has been generated, the recurrent decoder can employ it to generate the corresponding description. At every time-step of the decoding phase, the decoder unit uses the encoded vector $\mathbf{v}$, previous generated one-hot representation word $\textbf{w}_{t-1}$ and previous internal state $\mathbf{p}_{t-1}$ as input, and outputs a new internal state $\mathbf{p_{t}}$. Like~\cite{Baraldi2016}, our decoder unit is the Gated Recurrent Unit (GRU)~\cite{Cho2014}, a simplified version of LSTM, which is good at language decoding. The output of GRU is modulated via two sigmoid gates: a reset gate $\mathbf{r}_{t}$ and an update gate $\mathbf{z}_{t}$. The operation detail is as the following:
\begin{align}
\mathbf{z}_t &= \sigma(W_{zw} W_{w} \mathbf{w}_{t-1} + W_{zv} \mathbf{v} + W_{zp} \mathbf{p}_{t-1} + \mathbf{b}_z) \\
\mathbf{r}_t &= \sigma(W_{rw} W_{w} \mathbf{w}_{t-1} + W_{rv} \mathbf{v} + W_{rp} \mathbf{p}_{t-1} + \mathbf{b}_r).
\end{align}
Exploiting the values of the above gates, the output of the decoder at timestep $t$ is computed as:
\begin{align}
\tilde{\mathbf{p}}_t &= \phi(W_{pw} W_{w} \mathbf{w}_{t-1} + W_{pv} \mathbf{v} + W_{pp} (\mathbf{r}_t \odot \mathbf{p}_{t-1}) + \mathbf{b}_p)\\
\mathbf{p}_t &= (1-\mathbf{z}_t) \odot \mathbf{p}_{t-1} + \mathbf{z}_t \odot \tilde{\mathbf{p}}_t,
\end{align}
where $W_*$ and $\mathbf{b}_*$ are learned weights and biases and $W_w$ transforms the one-hot encoding of words to a dense lower dimensional embedding.
Again, $\odot$ denotes the element-wise product, $\sigma$ is the sigmoid function and $\phi$ is the hyperbolic tangent.
A softmax function is applied on $\mathbf{p}_{t}$ to compute the probability of producing certain word at current time-step:
\begin{equation}
p_{\omega}(\mathbf{w}_t | \mathbf{w}_{t-1}, \mathbf{w}_{t-2}, ..., \mathbf{w}_1, \mathbf{v}) = \mathbf{w}_t^T\text{softmax}(W_p \mathbf{p}_t),
\label{eq:wordp}
\end{equation}
where $W_p$ is used to project the output of the decoder to the dictionary space and $\omega$ denotes all parameters of the Encoder-Decoder. Also, the internal state $\mathbf{p}$ is initialized to zero. We use the greedy decode routine to generate every word. It means that at every time-step, we choose the word that has the maximal $p_{\omega}(\mathbf{w}_t | \mathbf{w}_{t-1}, \mathbf{w}_{t-2}, ..., \mathbf{w}_1, \mathbf{v})$ as the current output word. Specifically, we use a special token $<$BOS$>$ as $\textbf{w}_{0}$ to start the decoding, and when the decoder generates another special token $<$EOS$>$, the decoding procedure is terminated.

\begin{figure*}[!ht]
  \centering
  \includegraphics[width=\linewidth]{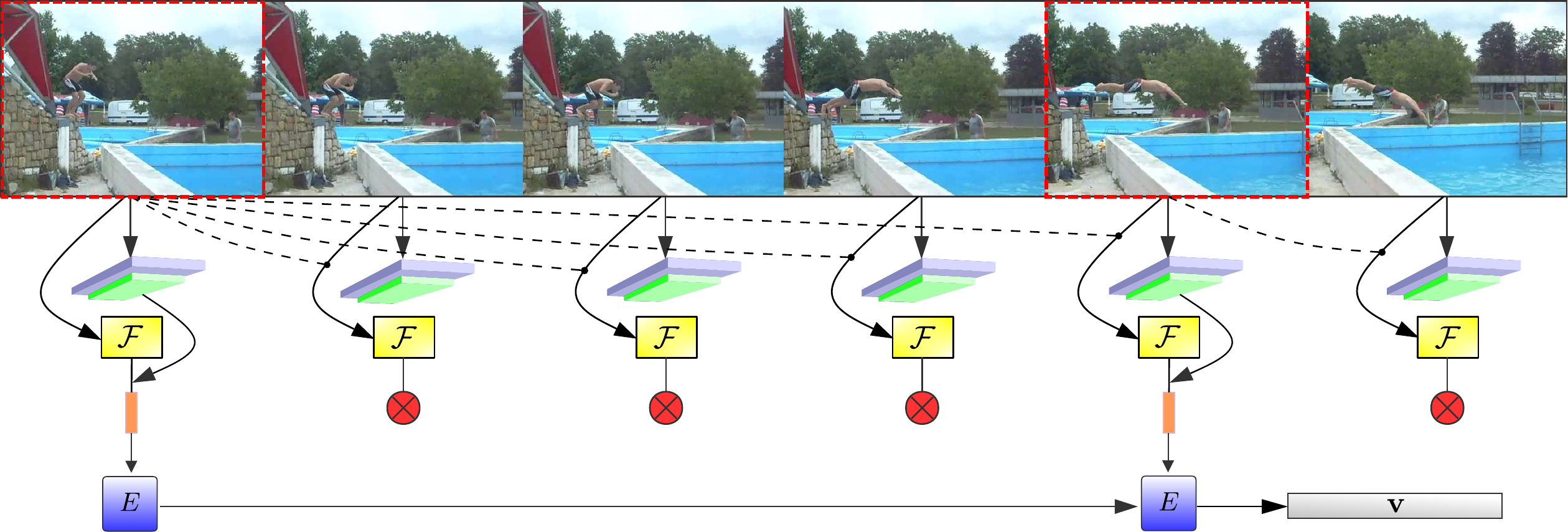}
\vspace{-1.5em}
\caption{A typical frame picking and encoding procedure of our framework. $\mathcal{F}$ denotes \textsf{PickNet}. $E$ is the encoder unit and $\mathbf{v}$ is the encoded video representation. The design choice is the balance between processing time and computation cost. The system can simultaneously extract convolutional features and decide whether to pick the frame or not at each time-step. If it decides not to pick the frame at certain time-step, the convolutional neural network can stop early to save computation cost. }
\label{fig:pickencoder}
\end{figure*}

\subsection{Our approach}
\subsubsection{Architecture}
\begin{figure}[H]
  \centering
  \includegraphics[width=\linewidth]{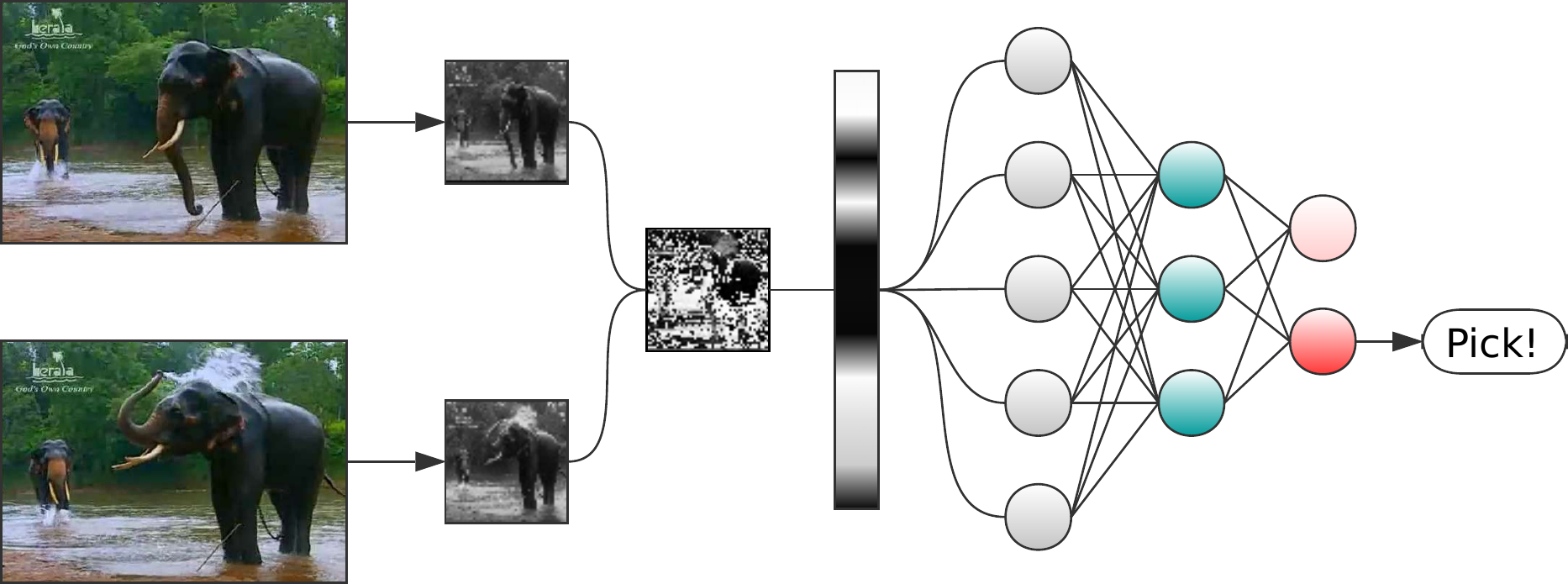}
\vspace{-1em}
\caption{The \textsf{PickNet} uses the flattened difference gray-scale image as input and produces a binomial distribution to indicate picking the current frame or not.}
\label{fig:pickent}
\end{figure}

The \textsf{PickNet} aims to select informative video content without knowing the global information. It means that the pick decision can only be based on the current observation and the history, which makes it more difficult than video summarization tasks. The more challenging thing is, we do not have supervised information to guide the learning of \textsf{PickNet} in video captioning tasks. Therefore, we formulate the problem as a reinforcement learning task, \ie, given an input image sequence sampled from a video, the agent should select a subset of them under certain policy to retain video content as much as possible. Here, we use \textsf{PickNet} to produce the picking policy. Figure~\ref{fig:pickent} shows the architecture of \textsf{PickNet}.
For the consideration of computation efficiency, we use a simple two-layer feedforward neural network as the prototype of \textsf{PickNet}. The network has two outputs, which indicate the probabilities to pick or drop the current observed frame.  We model the frame picking process as the glance-and-compare fashion. For each input frame $\mathbf{z}_{t}$, we first convert the colored image into grayscale image, and then resize it into a smaller image $\mathbf{g}_{t}$, which can be viewed as a ``glance'' of current frame. Then we subtract the current glance $\mathbf{g}_{t}$ by the glance of the last picked frame $\tilde{\mathbf{g}}$, to get a grayscale difference image $\mathbf{d}_{t}$; this can be seen as the ``compare''. Finally we flatten the 2D grayscale difference image into a 1D fixed size vector, and feed it to \textsf{PickNet} to produce a binomial distribution that the pick decision is sampled from:
\begin{equation}
  \mathbf{s}_{t} = W_{2}\cdot(\max(W_{1}\cdot\text{vec}(\mathbf{d}_{t})+\mathbf{b}_{1}, \mathbf{0}))+\mathbf{b}_{2}
\end{equation}
\begin{equation}
  p_{\theta}(a_{t}|\mathbf{z}_{t}, \tilde{\mathbf{g}}) \sim \text{softmax}(\mathbf{s}_{t}),
  \label{eq:policy}
\end{equation}
where $W_{*}$ are learned weight matrices and $\mathbf{b}_*$ are learned biases vectors.
During training, we use stochastic policy, i.e., the action is sampled according to Equation \eqref{eq:policy}. When testing, the policy becomes determined, hence the action with higher probability is chosen.
If the policy decides to pick the current frame, the frame feature will be extracted by a pretrained CNN and embedded into a lower dimension, then passed to the encoder unit, and the template will be updated:
\begin{equation}
  \tilde{\mathbf{g}} \leftarrow \mathbf{g}_{t}.
\end{equation}
We force \textsf{PickNet} to pick the first frame, thus the encoder will always process at least one frame, which makes the training procedure more robust. Figure~\ref{fig:pickencoder} shows how \textsf{PickNet} works with the encoder. It is worth noting that the input of \textsf{PickNet} can be of any other forms, such as the difference between optical flow maps, which may handle the motion information more properly.

\subsubsection{Rewards}
The design of rewards is very essential to reinforcement learning. For the purpose of picking informative video frames, we consider two parts of reward: the language reward and visual diversity reward.

\newcommand{\mathcider}{\text{CIDEr}\xspace}
\newcommand{\rs}{s\xspace}
\newcommand{\RS}{S\xspace}
\newcommand{\ngram}{$n$-gram\xspace}
\newcommand{\ngrams}{$n$-grams\xspace}
\newcommand{\cidern}{CIDEr$_n$\xspace}
\noindent \textbf{Language reward.} First of all, the picked frames should contain rich semantic information, which can be used to effectively generate language description. In the video captioning task, it is natural to use the evaluated language metrics as the language reward. Here, we choose CIDEr~\cite{Vedantam2015} score. Given a video $v_{i}$ and a collection of human generated reference sentences $\RS_i=\{s_{ij}\}$, the goal of CIDEr is to measure the similarity of the machine generated sentence $c_{i}$ to a majority of how most people describe the video. So the language reward $r_{l}$ is defined as:
\begin{equation}
  r_{l}(v_{i}, \RS_{i}) = \text{CIDEr}(c_{i}, \RS_{i})
\end{equation}

\noindent \textbf{Visual diversity reward.} Also, we want the picked frames that have good diversity in visual features. Using only language reward may miss some important visual information, so we introduce the visual diversity reward $r_{v}$. For all the selected frame features $\{\mathbf{x}_{i}\in \mathbb{R}^{D}\}$, we use the standard deviation of feature vectors as the visual diversity reward:
\begin{equation}
  r_{v}(v_{i}) = \sum_{j=1}^{D}\sqrt{\frac{1}{N_{p}}\sum_{i=1}^{N_{p}}(\mathbf{x}_{i}^{(j)}-\mu^{(j)})^{2}},
\end{equation}where $N_{p}$ is the number of picked frames, $\mathbf{x}_{i}^{(j)}$ is the $j$-th value of the $i$-th visual feature, and $\mu^{(j)}=\frac{1}{N_{p}}\sum_{i=1}^{N_{p}}\mathbf{x}_{i}^{(j)}$ is the mean of all the $j$-th value of visual features.

\noindent \textbf{Picks limitation.} If the number of picked frames is too large or too small, it may lead to poor performances in either efficiency or effectiveness, so we assign a negative reward to discourage this situations. Empirically, we set the minimum picked number $N_{\text{min}}$ as 3, which stands for beginning, highlight and ending. The maximum picked number $N_{\text{max}}$ is initially set as the $\frac{1}{3}$ of total frame number, and will be shrunk down along with the training process, until decreased to a minimum value $\tau$.

In summary, we merge the two parts of reward, and the final reward can be written as
\begin{equation}
  r(v_{i}) =
  \begin{cases}
    \lambda_{l}r_{l}(v_{i}, \RS_{i}) + \lambda_{v}r_{v}(v_{i}) & \text{if}\quad N_{\text{min}} \leq N_{p} \leq N_{\text{max}} \\
    R^{-} & \text{otherwise},
  \end{cases}
\end{equation}
where $\lambda_{*}$ is the weighting hyper-parameters and $R^{-}$ is the penalty reward.

\subsection{Training}
The training procedure is splitted into three stages. The first stage is to pretrain the Encoder-Decoder. We call it \textit{supervision} stage. In the second stage, we fix the Encoder-Decoder and train \textsf{PickNet} by reinforcement learning. It is called \textit{reinforcement} stage. And the final stage is the joint training of \textsf{PickNet} and the Encoder-Decoder. We call it \textit{adaptation} stage. We use standard back-propagation to train the Encoder-Decoder, and REINFORCE~\cite{Williams1992} to train \textsf{PickNet}.

\noindent \textbf{Supervision stage.} When training the Encoder-Decoder, traditional method maximizes the likelihood of the next ground-truth word given previous ground-truth words using back-propagation. However, this approach causes the \textit{exposure bias}~\cite{Ranzato2015}, which results in error accumulation during generation at test time, since the model has never been exposed to its own predictions. In order to alleviate this phenomenon, the schedule sampling~\cite{Bengio2015} procedure is used, which feeds back the model's own predictions and slowly increases the feedback probability during training. We use SGD with cross entropy loss to train the Encoder-Decoder. Given the ground-truth sentences $\mathbf{y}=(\mathbf{y}_{1}, \mathbf{y}_{2}, \ldots, \mathbf{y}_{m})$, the loss is defined as:
\begin{equation}
L_{\text{X}}(\omega)=-\sum_{t=1}^{m} \log (p_{\omega}(\mathbf{y}_{t}|\mathbf{y}_{t-1},\mathbf{y}_{t-2},\dots \mathbf{y}_{1}, \mathbf{v})),
\label{eq:xent}
\end{equation}
where $p_{\omega}(\mathbf{y}_{t}|\mathbf{y}_{t-1},\mathbf{y}_{t-2},\dots \mathbf{y}_{1}, \mathbf{v})$ is given by the parametric model in Equation \eqref{eq:wordp}.

\noindent \textbf{Reinforcement stage.} In this stage, we fix the Encoder-Decoder and treat it as the \textit{environment}, which can produce language reward to reinforce \textsf{PickNet}. The goal of training is to minimize the negative expected reward:
\begin{equation}
L_{R}(\theta)= -  \mathbb{E}_{\mathbf{a}^s\sim p_{\theta}} \left[ r(\mathbf{a}^s)\right],
\end{equation}
where $\theta$ denotes all parameters of \textsf{PickNet}, $p_{\theta}$ is the learned policy parameterized by Equation \eqref{eq:policy}, and $\mathbf{a}^s= (a^{s}_1, a^{s}_{2}, \ldots, a^{s}_n)$ while $a^{s}_{t}$ is the action sampled from the learned policy at the time step $t$. 

We train \textsf{PickNet} by using REINFORCE algorithm, which is based on the observation that the gradient of a non-differentiable expected reward can be computed as follows:
\begin{equation}
\nabla_{\theta}L_{R}(\theta)= - \mathbb{E}_{\mathbf{a}^s \sim p_{\theta}} \left[ r(\mathbf{a}^s) \nabla_{\theta} \log p_{\theta}(\mathbf{a}^s) \right].
\end{equation}
Using the chain rule, the gradient can be rewritten as:
\begin{align}
  \nabla_{\theta}L_{R}(\theta)= & \sum_{t=1}^{n} \frac{\partial L_{R}(\theta)}{\partial \mathbf{s}_{t}}\frac{\partial \mathbf{s}_{t}}{\partial \theta} \\
   = & \sum_{t=1}^{n}- \mathbb{E}_{\mathbf{a}^s \sim p_{\theta}}r(\mathbf{a}^s)(p_{\theta}(a^{s}_{t})- \mathbf{1}_{a^{s}_{t}})\frac{\partial \mathbf{s}_{t}}{\partial \theta},
\end{align}
where $\mathbf{s}_{t}$ is the input to the softmax function. In practice, the gradient can be approximated using a single Monte-Carlo sample $\mathbf{a}^s=(a^{s}_{1}, a^{s}_{2}, \ldots, a^{s}_{n})$ from $p_{\theta}$:
\begin{equation}
\nabla_{\theta}L_{R}(\theta)\approx -\sum_{t=1}^{n}r(\mathbf{a}^s)(p_{\theta}(a^{s}_{t})- \mathbf{1}_{a^{s}_{t}})\frac{\partial \mathbf{s}_{t}}{\partial \theta}.
\end{equation}
When using REINFORCE to train the policy network, we need to estimate a baseline reward $b$ to diminish the variance of gradients. Here, the \textit{self-critical}~\cite{Rennie2016} strategy is used to estimate $b$. In brief, the reward obtained by current model under inferencing used at test stage, denoted as $r(\hat{\mathbf{a}})$, is treated as the baseline reward. Therefore, the final gradient expression is:
\begin{equation}
\nabla_{\theta}L_{R}(\theta)\approx -(r(\mathbf{a}^s)-r(\hat{\mathbf{a}}))\sum_{t=1}^{n}(p_{\theta}(a^{s}_{t})- \mathbf{1}_{a^{s}_{t}})\frac{\partial \mathbf{s}_{t}}{\partial \theta}.
\end{equation}

\noindent \textbf{Adaptation stage.} After the first two stages, the Encoder-Decoder and \textsf{PickNet} are well pretrained, but there exists a gap between them because the Encoder-Decoder use the full video frames as input while \textsf{PickNet} just selects a portion of frames. So we need a joint training stage to integrate this two parts together. However, the pick action is not differentiable, so the gradients introduced by cross-entropy loss can not flow into \textsf{PickNet}. Hence, we follow the approximate joint training scheme. 
In each iteration, the forward pass generates frame picks which are treated just like fixed picks when training the Encoder-Decoder, and the backward propagation and REINFORCE updates are performed as usual. It acts like performing dropout in time sequence, which can improve the versatility of the Encoder-Decoder.
\vspace{-1ex}

\section{Experimental Setup}
\vspace{-1ex}
\subsection{Datasets}
We evaluate our model on two widely used video captioning benchmark datasets: the Microsoft Video Description (MSVD)~\cite{Chen2011a} and the MSR Video-to-Text (MSR-VTT)~\cite{Xu2016}.

\noindent \textbf{Microsoft Video Description (MSVD).} The Microsoft Video Description is also known as YoutubeClips. It contains 1,970 Youtube video clips, each labeled with around 40 English descriptions collected by Amazon Mechanical Turkers. As done in previous works~\cite{Venugopalan2015}, we split the dataset into three parts: the first 1,200 videos for training, then the followed 100 videos for validation and the reset 670 videos for test. This dataset mainly contains short video clips with a single action, and the average duration is about 9 seconds. So it is very suitable to use only a portion of frames to represent the full video.

\noindent \textbf{MSR Video-to-Text (MSR-VTT).} The MSR Video-to-Text is a large-scale benchmark for video captioning. It provides 10,000 video clips, and each video is annotated with 20 English descriptions. Thus, there are 200,000 video-caption pairs in total. This dataset is collected from a commercial video search engining and so far it covers the most comprehensive categories and diverse visual contents. Following the original paper, we split the dataset in contiguous groups of videos by index number: 6,513 for training, 497 for validation and 2,990 for test.

\begin{figure*}[th]
  \subfigure{
    \begin{minipage}[t]{0.5\linewidth}
      \vspace{0pt}
      \begin{center}
        \includegraphics[width=\columnwidth]{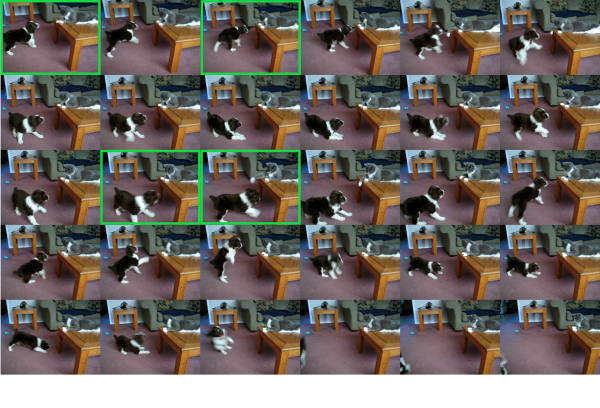} \\
        \vspace{-1.4em}
        \small{
          \textbf{Ours: a cat is playing with a dog} \\
          \textbf{GT: a dog is playing with a cat}
        }
      \end{center}
    \end{minipage}
  }
  \subfigure{
    \begin{minipage}[t]{0.5\linewidth}
      \vspace{0pt}
      \begin{center}
        \includegraphics[width=\columnwidth]{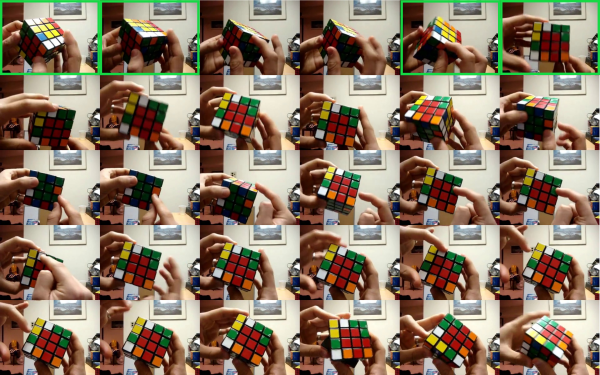}
        \small{
          \textbf{Ours: a person is solving a rubik's cube} \\
          \textbf{GT: person playing with toy}
        }
      \end{center}
    \end{minipage}
  }
\vspace{-1em}
\caption{Example results on MSVD (left) and MSR-VTT (right). The green boxes indicate picked frames. (Best viewed in color and zoom-in. Frames are organized from left to right, then top to bottom in temporal order. )}
\label{fig:Result}
\end{figure*}

\subsection{Metrics}
We employ four popular metrics for evaluation: BLEU~\cite{Papineni2002}, ROUGE$_{L}$~\cite{Lin2004}, METEOR~\cite{Banerjee2005} and CIDEr. As done in previous video captioning works, we use METEOR and CIDEr as the main comparison metrics. In addition, Microsoft COCO evaluation server~\cite{Chen2015a} has implemented these metrics and release evaluation functions\footnote{https://github.com/tylin/coco-caption}, so we directly call such evaluation functions to test the performance of video captioning. Also, the CIDEr reward is computed by these functions.

\subsection{Video preprocessing}
First, we sample equally-spaced 30 frames for every video, and resize them into 224$\times$224 resolution. Then the images are encoded with the final convolutional layer of ResNet152~\cite{He2015}, which results in a 2,048-dimensional vector. Most video captioning models use motion features to improve performance. However, we only use the appearance features in our model, because extracting motion features is very time-consuming, which deviates from our purpose that cutting down the computation cost for video captioning.

\subsection{Text preprocessing}
We tokenize the labeled sentences by converting all words to lowercases and then utilizing the word\_tokenize function from NLTK\footnote{http://www.nltk.org/} toolbox to split sentences into words and remove punctuation. Then, the word with frequency less than 3 is removed. As a result, we obtain the vocabulary with 5,491 words from MSVD and 13,065 words from MSR-VTT. For each dataset, we use the one-hot vector (1-of-$N$ encoding, where $N$ is the size of vocabulary) to represent each word.

\subsection{Implementation details}
We use the validation set to tune some hyperparameters of our framework. The learning rates for three training stages are set to $3\times 10^{-4}$, $3\times 10^{-4}$ and $1\times 10^{-4}$, respectively. The training batchsize is 128 for MSVD and 256 for MSR-VTT, while each stage is trained up to 100 epoches and the best model is used to initialize the next stage. The minimum value of maximum picked frames $\tau$ is set to 7, and the penalty reward $R^{-}$ is $-1$.  To regularize the training and avoid over-fitting, we apply the well known regularization technique Dropout with retain probability 0.5 on the input and output of the encoding LSTMs and decoding GRUs. Embeddings for video features and words have size 512, while the sizes of all recurrent hidden states are empirically set to 1,024. For \textsf{PickNet}, the size of glance is 56$\times$56, and the size of hidden layer is 1,024. The Adam~\cite{Kingma2015a} optimizer is used to update all the parameters.

\section{Results and Discussion}
\vspace{-1ex}
Figure~\ref{fig:Result} gives some example results on the test sets of two datasets. As it can be seen, our \textsf{PickNet} can select informative frames, so the rest of our model can use these selected frames to generate reasonable descriptions. More results are offered in supplemental materials. In order to demonstrate the effectiveness of our framework, we compare our approach with some state-of-the-art methods on the two datasets, and analyze the learned picks of \textsf{PickNet}.

\subsection{Comparison with the state-of-the-arts}
We compare our approach on MSVD with six state-of-the-art approaches for video captioning: TA~\cite{Yao2015a}, S2VT~\cite{Venugopalan2015}, LSTM-E~\cite{Pan2016c}, p-RNN~\cite{Yu2016a} HRNE~\cite{Pan2016d} and BA~\cite{Baraldi2016}. \mbox{LSTM-E} uses a visual-semantic embedding to generate better captions. TA uses temporal attention while \mbox{p-RNN} use both temporal and spatial attention. BA uses a hierarchical encoder while HRNE use a hierarchical decoder to describe videos. S2VT uses stack LSTMs both for the encode and decode stage. All of these methods use motion features (C3D or optical flow) and extract visual features frame by frame. Besides, we report the performance of our baseline model, which encodes all the sampled frames. In order to compare our \textsf{PickNet} with other picking policies, we conduct two other trials that pick frames by randomly selecting and $k$-means clustering, respectively. Also, for analyzing the effect of different rewards, we conduct some ablation studies on them. As it can be noticed in Table~\ref{tab:msvd}, our method improves plain techniques and achieves the state-of-the-art performance on MSVD. This result outperforms the most recent state-of-the-art method by a considerable margin of $\frac{76.0-65.8}{65.8}\approx 15.5\%$ on the CIDEr metric. Further, we try to compare the time efficiency among these approaches. However, most of state-of-the-art methods do not release executable codes, so the accurate performance may not be available. Instead, we estimate the running time by the complexity of visual feature extractors and the number of processed frames. The details of running time estimation are listed in supplemental materials. Thanks to the \textsf{PickNet}, our captioning model is $4 \scriptsize{\sim}33$ times faster than other methods.

\begin{table}[tb]
\begin{center}
\resizebox{\columnwidth}{!}{
\begin{tabular}{l|c|c|c|c|r}
  \hline
  Model                              & BLEU@4 & ROUGE-L & METEOR & CIDEr & Time \\
  \hline
  \multicolumn{6}{c}{Previous Work} \\
  \hline
  TA\cite{Yao2015a}                  & 41.9   & -       & 29.6   & 51.7  & 4x \\
  S2VT\cite{Venugopalan2015}         & -      & -       & 29.8   & -     & 13x \\
  LSTM-E\cite{Pan2016c}              & 45.3   & -       & 31.0   & -     & 5x \\
  p-RNN\cite{Yu2016a}                & \textbf{49.9}   & -       & 32.6  & 65.8  & 5x \\
  HRNE\cite{Pan2016d}                & 43.8   & -       & \textbf{33.1}  & -     & 33x \\
  BA\cite{Baraldi2016}               & 42.5   & -       & 32.4   & 63.5  & 12x \\
  \hline
  \multicolumn{6}{c}{Our Models} \\
  \hline
  Baseline                           & 44.8   & 68.5    & 31.6   & 69.4  & 5x \\
  Random                             & 35.6   & 64.5    & 28.4   & 49.2  & 2.5x \\
  $k$-means ($k$=6)                      & 45.2   & 68.5    & 32.4   & 70.9  & 1x \\
  \textsf{PickNet} (V)               & 43.6   & 68.4    & 32.4   & 75.6  & 1x \\
  \textsf{PickNet} (L)               & \textbf{49.9}    & \textbf{69.3}  & 32.9   & 74.7 & 1x\\ \hline
  \textsf{PickNet} (V+L)             & 46.1   &  69.2   & \textbf{33.1}  & \textbf{76.0} & 1x\\ \hline
\end{tabular}
}
\end{center}
\vspace{-1em}
\caption{Experiment results on MSVD. L denotes using language reward and V denotes using visual diversity reward. $k$ is set to the average number of picks $\bar{N_{p}}$ on MSVD. ($\bar{N_{p}}\approx 6$)}
\vspace{-1.5em}
\label{tab:msvd}
\end{table}

On MSR-VTT, we compare four state-of-the-art approaches: ruc-uva~\cite{Dong2016}, Aalto~\cite{Shetty2016}, DenseCap~\cite{Shen2017} and \mbox{MS-RNN}~\cite{Song2017b}. \mbox{ruc-uva} incorporates the Encoder-Decoder with two new stages called early embedding which enriches input with tag embeddings, and late reranking which re-score generated sentences in terms of their relevance to a specific video. Aalto first trains two models which are separately based on attribute and motion features, and then trains a evaluator to choose the best candidate generated by the two captioning model. DenseCap generates multiple sentences with regard to video segments and uses a winner-take-all scheme to produce the final description. \mbox{MS-RNN} uses a multi-modal LSTM to model the uncertainty in videos to generate diverse captions. Compared with these methods, our method can be simply trained in end-to-end fashion, and does not rely upon any attribute information. The performance of these approaches and that of our solution is reported in Table~\ref{tab:msrvtt}. We observe that our approach is able to achieve competitive result even without utilizing attribute information, while other methods take advantage of attributes and auxiliary information sources. Also, our model is fastest. It is also worth noting that the \textsf{PickNet} can be easily integrated with the compared methods, since none of them incorporated with frame selection algorithm. For example, DenseCap generates region-sequence candidates based on equally sampled frames. It can alternatively utilize \textsf{PickNet} to reduce the time for generating candidates by cutting down the number of selected frames.

\begin{table}[tb]
\begin{center}
\resizebox{\columnwidth}{!}{
\begin{tabular}{l|c|c|c|c|r}
  \hline
  Model                              & BLEU@4 & ROUGE-L & METEOR & CIDEr & Time \\
  \hline
  \multicolumn{6}{c}{Previous Work} \\
  \hline
  ruc-uva~\cite{Dong2016}            & 38.7   & 58.7    & 26.9   & 45.9  & 4.5x \\
  Aalto~\cite{Shetty2016}            & 39.8   & 59.8    & 26.9   & 45.7  & 4.5x \\
  DenseCap~\cite{Shen2017}           & 41.4   & 61.1    & 28.3   & 48.9  & 3.5x \\
  MS-RNN~\cite{Song2017b}            & 39.8   & 59.3    & 26.1   & 40.9  & 10x  \\
  \hline
  \multicolumn{6}{c}{Our Models} \\
  \hline
  Baseline                           & 36.8   & 59.0    & 26.7   & 41.2  & 3.8x \\
  Random                             & 31.3   & 55.7    & 25.2   & 32.6  & 1.9x \\
  $k$-means ($k$=8)                  & 37.8   & 59.1    & 26.9   & 41.4  & 1x \\
  \textsf{PickNet} (V)               & 35.6   & 58.2    & 26.8   & 41.0  & 1x \\
  \textsf{PickNet} (L)               & 37.3   & 58.9    & 27.0   & 41.9  & 1x \\ \hline
  \textsf{PickNet} (V+L)             & {\color{blue}38.9}   & {\color{blue}59.5}    & {\color{blue}27.2}   & {\color{blue}42.1}  & 1x \\ \hline
\end{tabular}
}
\end{center}
\vspace{-1em}
\caption{Experiment results on MSR-VTT. $k$ is set to the average number of picks $\bar{N_{p}}$ on MSR-VTT. ($\bar{N_{p}}\approx 8$)}
\vspace{-1.5em}
\label{tab:msrvtt}
\end{table}

\subsection{Analysis of learned picks}
\begin{figure}[!bt]
  \subfigure[Dist. of the number of picks.\label{fig:PickAnalysisNum}]{
    \begin{minipage}[t]{0.47\linewidth}
      \vspace{0pt}
      \begin{center}
        \includegraphics[width=\columnwidth]{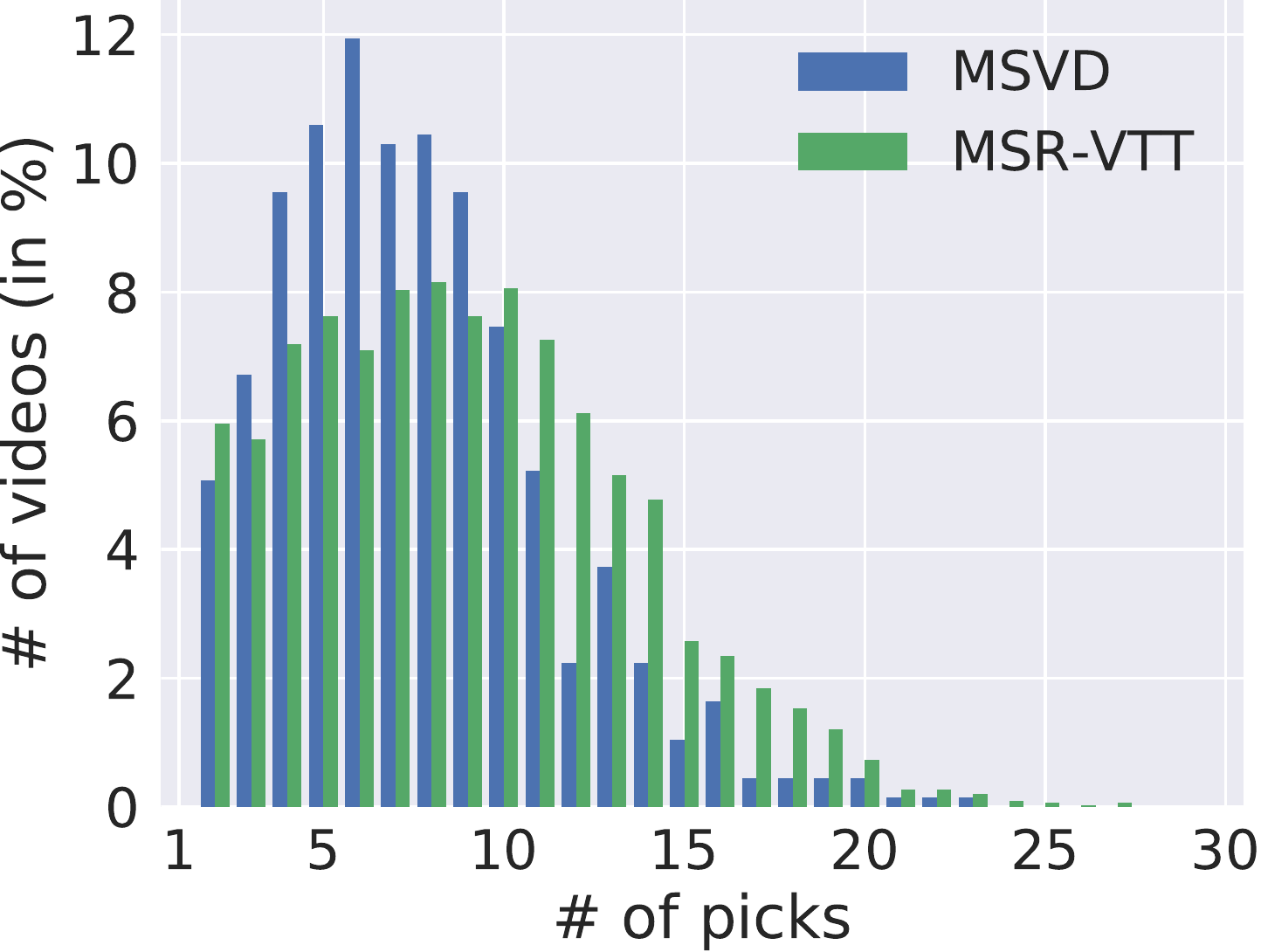}
      \end{center}
    \end{minipage}
  }
  \subfigure[Dist. of the position of picks.\label{fig:PickAnalysisPosition}]{
    \begin{minipage}[t]{0.47\linewidth}
      \vspace{0pt}
      \begin{center}
        \includegraphics[width=\columnwidth]{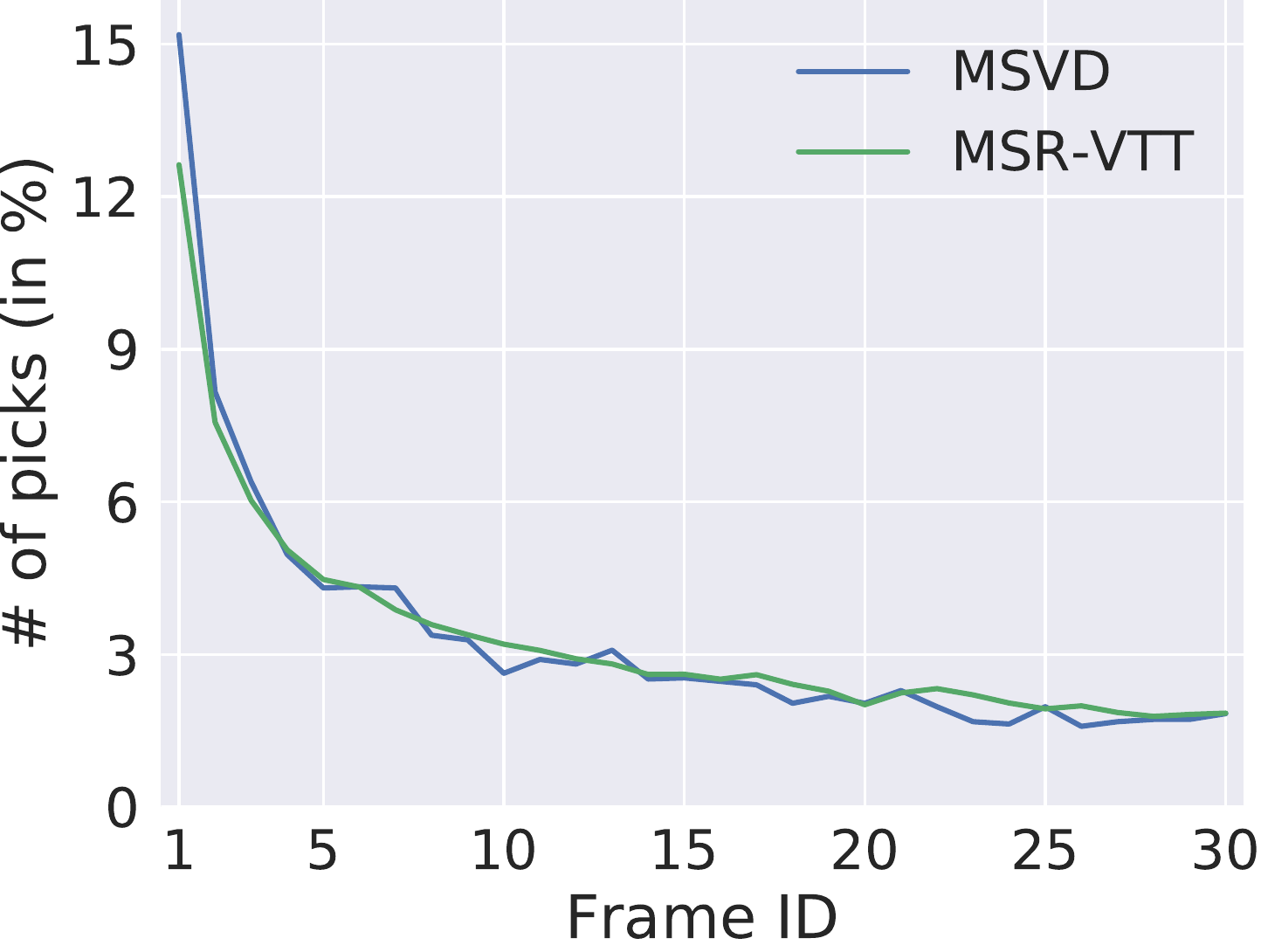}
      \end{center}
    \end{minipage}
  }
\vspace{-1em}
\caption{Statistics on the behavior of our \textsf{PickNet}.}
\vspace{-1em}
\label{fig:PickAnalysis}
\end{figure}
We collect statistics on the properties of our \textsf{PickNet}. Figure~\ref{fig:PickAnalysis} shows the distributions of the number and position of picked frames on the test sets of MSVD and MSR-VTT. As observed in Figure~\ref{fig:PickAnalysisNum}, in the vast majority of the videos, less than 10 frames are picked. It implies that in most case only $\frac{10}{30}\approx 33.3\%$ frames are necessary to be encoded for captioning videos, which can largely reduce the computation cost. Specifically, the average number of picks is around $6$ for MSVD and $8$ for MSR-VTT. Looking at the distributions of position of picks in Figure~\ref{fig:PickAnalysisPosition}, we observe a pattern of \textit{power law distribution}, \ie, the probability of picking a frame is reduced as time goes by. It is reasonable since most videos are single-shot and the anterior frames are sufficient to represent the whole video.

\subsection{Captioning for streaming video}
\begin{figure}[htb]
  \centering
  \includegraphics[width=\linewidth]{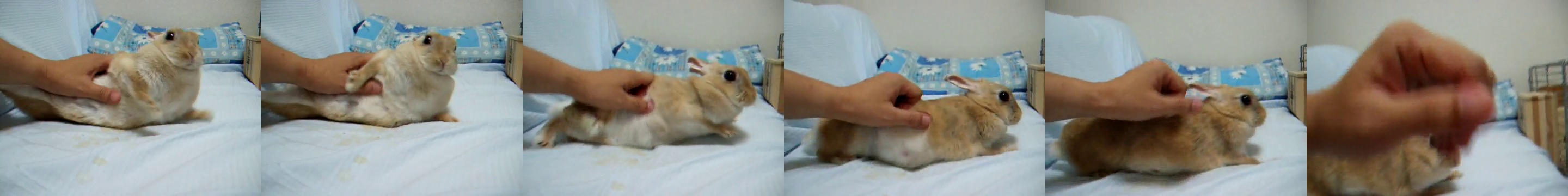}
  {\footnotesize \textbf{a cat is playing $\rightarrow$ a rabbit is playing $\rightarrow$ a rabbit is being petted \\ $\rightarrow$  {\color{blue} a person is petting a rabbit} $\times 3$}}
\vspace{-0.5em}
\caption{An example of online video captioning.}
\vspace{-1em}
\label{fig:online}
\end{figure}

One of the advantage of our method is that it can be applied to streaming video. Different from offline video captioning, captioning for streaming video requires the model to tackle with unbounded video and generate descriptions immediately when the visual information has changed, which meets the demand of practical applications. For this online setting, we first sample frames at 1fps, and then sequentially feed the sampled frames to \textsf{PickNet}. If certain frame is picked, the pretrained CNN will be used to extract visual features of this frame. After that, the encoder will receive this feature, and produce a new encoded representation of the video stream up to current time. Finally, the decoder will generate a description based on the encoded representation. Figure~\ref{fig:online} demonstrates an example of online video captioning with the picked frames and corresponding descriptions. As it is shown, the descriptions will be more appropriate and more determined as the informative frames are picked. More results can be seen in supplemental materials.

\section{Conclusion}
\vspace{-1ex}
In this work, we design a plug-and-play reinforcement-learning-based \textsf{PickNet}  to select informative frames for the task of video captioning, which achieves  promising performance on effectiveness, efficiency and flexibility on popular benchmarks. This architecture can largely cut down the usage of convolution operations by picking only $6 \scriptsize{\sim}8$ frames for a video clip, while other video analysis methods usually require more than 40 frames. This property makes our method more applicable for real-world video processing. The proposed \textsf{PickNet} has a good flexibility and could be potentially employed to other video-related applications, such as video classification and action detection, which will be further addressed in our future work.

{\small
\bibliographystyle{ieee}
\bibliography{cited}
}

\newpage
\clearpage
\onecolumn
\title{Supplemental Materials}
\author{}
\date{}
\maketitle

\vspace{-3em}
\section{Details on Time Estimation}
\begin{table}
\begin{center}
\begin{tabular}{l|c|c|l|c|r}
  \hline
  Model                              & Appearance & Motion & Sampling method & Frame num. & Time \\
  \hline
  \multicolumn{6}{c}{Previous Work} \\
  \hline
  TA\cite{Yao2015a}                  & GoogleNet (0.5x) & C3D (2x)  & uniform sampling 26 frames  & 26 (4x)   & $0.5\times 2 \times 4 = 4$x \\
  S2VT\cite{Venugopalan2015}         & VGG (0.5x)       & OF (2x)   & uniform sampling 80 frames  & 80 (13x)  & $0.5\times 2\times 13 = 13$x \\
  LSTM-E\cite{Pan2016c}              & VGG (0.5x)       & C3D (2x)  & uniform sampling 30 frames  & 30 (5x)   & $0.5\times 2\times 5=5$x \\
  \textit{p}-RNN\cite{Yu2016a}                & VGG (0.5x)       & C3D (2x)  & uniform sampling 30 frames  & 30 (5x)   & $0.5\times 2\times 5=5$x \\
  HRNE\cite{Pan2016d}                & GoogleNet (0.5x) & C3D (2x)  & first 200 frames  & 200 (33x) & $0.5\times 2\times 33=33$x \\
  BA\cite{Baraldi2016}               & ResNet (0.5x)    & C3D (2x)  & every 5 frames & 72 (12x)   & $0.5\times 2\times 12=12$x \\
  \hline
  \multicolumn{6}{c}{Our Models} \\
  \hline
  Baseline                           & ResNet (1x)  & $\times$  & uniform sampling 30 frames  & 30 (5x)   & $1\times 5 =5$x \\
  Random                             & ResNet (1x)  & $\times$  & randomly sampling  & 15 (2.5x) & $1\times 2.5 =2.5$x \\
  $k$-means ($k$=6)                  & ResNet (1x)  & $\times$  & $k$-means clustering  & 6 (1x)    & $1\times 1=1$x \\
  \textsf{PickNet} (V)               & ResNet (1x)  & $\times$  & picking  & 6 (1x)    & $1\times 1=1$x \\
  \textsf{PickNet} (L)               & ResNet (1x)  & $\times$  & picking  & 6 (1x)    & $1\times 1=1$x\\ \hline
  \textsf{PickNet} (V+L)             & ResNet (1x)  & $\times$  & picking  & 6 (1x)    & $1\times 1=1$x\\ \hline
\end{tabular}
\end{center}
\caption{Running time estimation on MSVD. OF means optical flow. BA uses ResNet50 while our models use ResNet152. $k$ is set to the average number of picks $\bar{N_{p}}$ on MSVD. ($\bar{N_{p}}\approx 6$)}
\label{tab:msvd}
\end{table}

\begin{table}
\begin{center}
\begin{tabular}{l|c|c|l|c|r}
  \hline
  Model                              & Appearance & Motion & Sampling method & Frame num. & Time \\
  \hline
  \multicolumn{6}{c}{Previous Work} \\
  \hline
  ruc-uva~\cite{Dong2016}            & GoogleNet (0.5x) & C3D (2x) & every 10 frames & 36 (4.5x)  & $0.5\times 2\times 4.5=4.5$x \\
  Aalto~\cite{Shetty2016}            & GoogleNet (0.5x) & C3D+IDT (2x) & one frame every second & 36 (4.5x) & $0.5\times 2\times 4.5=4.5$x \\
  DenseCap~\cite{Shen2017}           & ResNet (0.5x)  & C3D (2x) & uniform sampling 30 frames  & 30 (3.5x)  & $0.5\times 2\times 3.5=3.5$x \\
  MS-RNN~\cite{Song2017b}            & ResNet (1x)   & C3D (2x) & uniform sampling 40 frames  & 40 (5x)  & $1\times 2\times 5=10$x  \\
  \hline
  \multicolumn{6}{c}{Our Models} \\
  \hline
  Baseline                           & ResNet (1x)  & $\times$  & uniform sampling 30 frames & 30 (3.8x)   & $1\times 3.8=3.8$x \\
  Random                             & ResNet (1x)  & $\times$  & randomly sampling & 15 (1.9x)   & $1\times 1.9=1.9$x \\
  $k$-means ($k$=8)                  & ResNet (1x)  & $\times$  & $k$-means clustering & 8 (1x)   & $1\times 1=1$x \\
  \textsf{PickNet} (V)               & ResNet (1x)  & $\times$  & picking &  8 (1x)   & $1\times 1=1$x \\
  \textsf{PickNet} (L)               & ResNet (1x)  & $\times$  & picking &  8 (1x)   & $1\times 1=1$x \\ \hline
  \textsf{PickNet} (V+L)             & ResNet (1x)  & $\times$  & picking &  8 (1x)   & $1\times 1=1$x \\ \hline
\end{tabular}
\end{center}
\vspace{-1.5em}
\caption{Running time estimation on MSR-VTT. IDT means improved dense trajectory. DenseCap uses ResNet50 while our models use ResNet152. $k$ is set to the average number of picks $\bar{N_{p}}$ on MSR-VTT. ($\bar{N_{p}}\approx 8$)}
\label{tab:msrvtt}
\end{table}
We estimate the running time by the complexity of visual feature extractors and the number of processed frames, and our \textsf{PickNet} (V+L) is treated as the baseline. For visual features, both appearance and motion are taken into consideration. The relative running time of different CNNs is based on public reports~\cite{cnn}. If a specific model uses motion features, the total computation time will be doubled, since extracting motion features is very time-consuming. For each model, the number of processed frames is set to the expected number of sampled frames under the sampling method. However, some model sample every 5 or 10 frames from input video, so the expected number depends on video length. In order to compare these model with others, we consider the input video with a duration of 10 seconds and a frame rate of 36fps, thus the total number of frames for input video is fixed to 360. Table~\ref{tab:msvd} and Table~\ref{tab:msrvtt} show the detail of time estimation on MSVD and MSR-VTT.

It is worth mentioning that the estimated time on other compared approaches is just rough. In fact, the actual speedup of our model will be higher than we estimated, because most of the compared models use complex pipeline. For instance, in the \textit{p}-RNN, the temporal- and spatial-attention mechanisms are exploited to selectively focus on visual elements during sentence generation. Also, in the DenseCap, a lexical FCN is utilized to extract region features of every frames, then the cost-effective lazy forward (CELF) method is employed to generate region-sequence candidates, and finally a bi-directional LSTM is used to encode each region-sequence candidate. All the above procedures are far more complex than our Encoder-Decoder pipeline, therefore will consume much more processing time.

\section{More Result Examples}
\begin{figure}
  \subfigure{
    \begin{minipage}[t]{0.5\linewidth}
      \vspace{0pt}
      \begin{center}
        \includegraphics[width=\columnwidth]{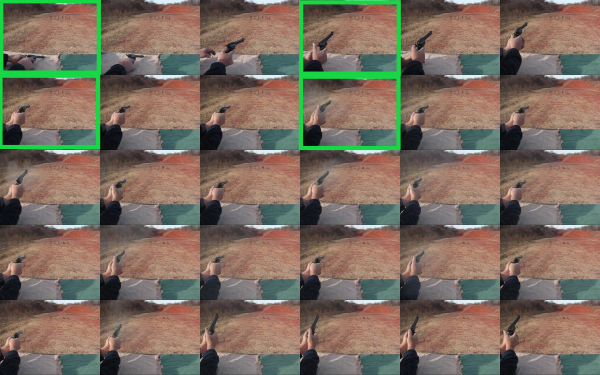}
        (a)
        \small{
                \textbf{Ours: a man is shooting a gun} \\
                \textbf{GT: a man is shooting}  
        }
      \end{center}
    \end{minipage}
  }
  \vspace{-0.5em}
  \subfigure{
    \begin{minipage}[t]{0.5\linewidth}
      \vspace{0pt}
      \begin{center}
        \includegraphics[width=\columnwidth]{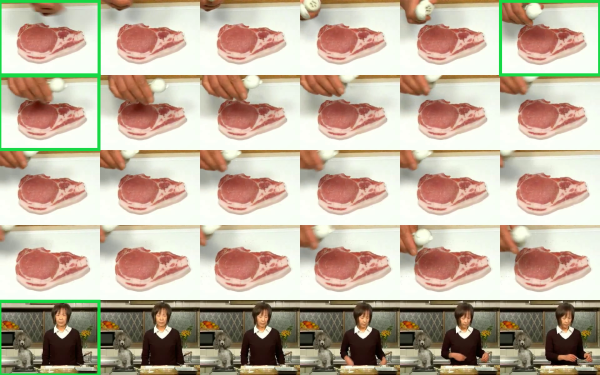} \\
        (b)
        \small{
                \textbf{Ours: a woman is seasoning meat} \\
                \textbf{\hspace{2.2em} GT: someone is seasoning meat}  
        }
      \end{center}
    \end{minipage}
  }
  \subfigure{
    \begin{minipage}[t]{0.5\linewidth}
      \vspace{0pt}
      \begin{center}
        \includegraphics[width=\columnwidth]{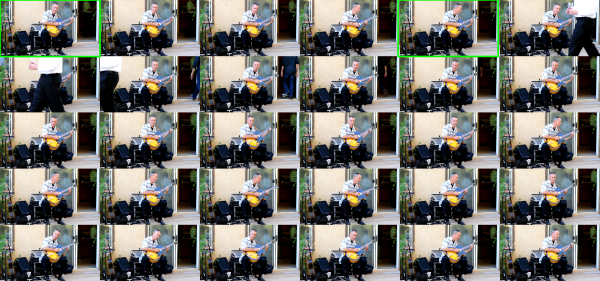} \\
        (c)
        \small{
                \textbf{Ours: a man is playing a guitar} \\
                \textbf{\hspace{2.3em} GT: a man is playing a guitar}  
        }
      \end{center}
    \end{minipage}
  }
  \vspace{-0.5em}
  \subfigure{
    \begin{minipage}[t]{0.5\linewidth}
      \vspace{0pt}
      \begin{center}
        \includegraphics[width=\columnwidth]{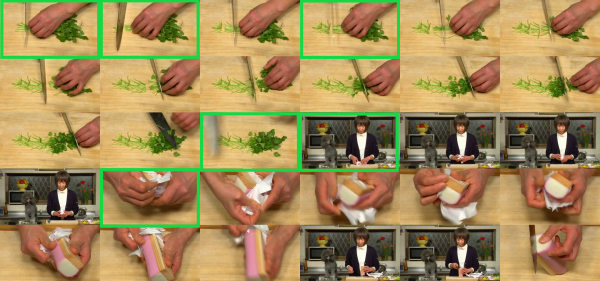}
        (d)
        \small{
                \textbf{Ours: a woman is chopping some leaves} \\
                \textbf{\hspace{1.1em} GT: a woman is slicing some leaves}  
        }
      \end{center}
    \end{minipage}
  }
  \subfigure{
    \begin{minipage}[t]{0.5\linewidth}
      \vspace{0pt}
      \begin{center}
        \includegraphics[width=\columnwidth]{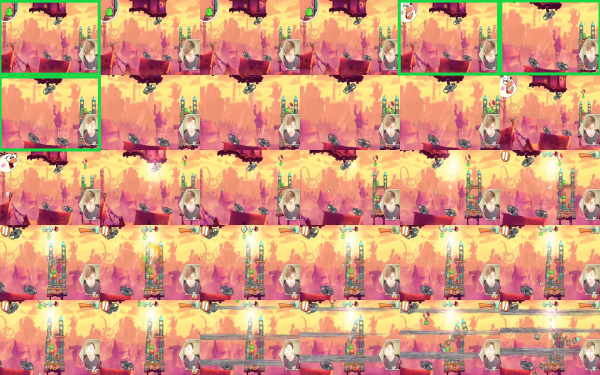} \\
        (e)
        \small{
                \textbf{Ours: a person is playing a video game} \\
                \textbf{\hspace{-1.9em} GT: a game is being played}  
        }
      \end{center}
    \end{minipage}
  }
  \subfigure{
    \begin{minipage}[t]{0.5\linewidth}
      \vspace{0pt}
      \begin{center}
        \includegraphics[width=\columnwidth]{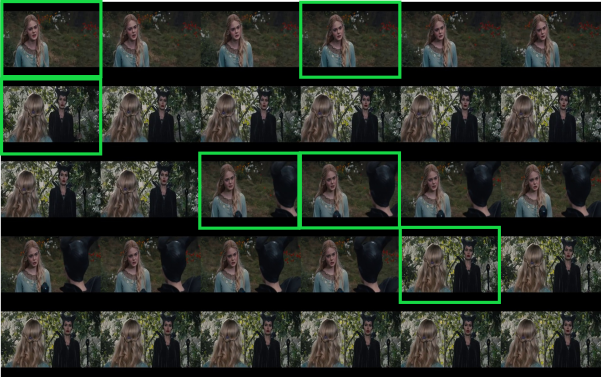}
        (f)
        \small{
                \textbf{Ours: there is a woman is talking with a woman} \\
                \textbf{\hspace{0em} GT: it is a movie}  
        }
      \end{center}
    \end{minipage}
  }
\caption{Example results on the test set of MSVD and MSR-VTT. The green boxes indicate picked frames. (Best viewed in color and zoom-in. Frames are organized from left to right, then top to bottom in temporal order. )}
\label{fig:MoreResult}
\end{figure}

\subsection{Offline video captioning}
Figure~\ref{fig:MoreResult} shows more results for offline video captioning on MSVD and MSR-VTT. As mentioned before, our \textsf{PickNet} can select a minor portion of frames to represent the whole video and describe the video with regard to picked frames. Moreover, in the left column, only 3 or 4 frames are picked, and these picked frames are all in the front part of the video. We suppose that it is because these videos contain univocal content, and in each video most frames are similar. Under these circumstances, it is enough to pick a few of frames at the beginning to describe these video. Meanwhile, in the right column, more frames are picked, and the picked frames are scattered. We suggest that it is because the content of these videos is diverse. In this situation, the \textsf{PickNet} will traverse the whole video and select more frames.

Altogether, two characteristics of picked frames can be found. \textbf{The first characteristic is that the picked frames are concise and highly related to the generated descriptions.} For example, in Figure~\ref{fig:MoreResult}(a), our model only selects four frames, which correspond to \textit{holding the gun}, \textit{porting arms}, \textit{aiming}, and \textit{shooting}, respectively. All other frames are more or less duplicated visually or semantically, so those redundant frames are ignored. In Figure~\ref{fig:MoreResult}(c), our model selects the 5th frame instead of the 6th frame. Although the 6th frame is more diverse than the 5th frame, it is not related to the description, so our model does not select it but pick the 5th frame to confirm that the clip is about \textit{playing a guitar}. In Figure~\ref{fig:MoreResult}(f), the picked frames appropriately describe the shot change, therefore the model can focus on the two women and understand they are talking to each other. \textbf{The second one is that the adjacent frames may be picked to represent action.} For example, in Figure~\ref{fig:MoreResult}(b), our model selects a pair of adjacent frames, i.e., the 6th and the 7th frames, which can properly represent the \textit{seasoning} action. In Figure~\ref{fig:MoreResult}(d), the first two frames are picked to represent the \textit{chopping} action. In Figure~\ref{fig:MoreResult}(e), the 5th to 7th frames are selected to represent the \textit{playing} action. And in Figure~\ref{fig:MoreResult}(f), the 15th and the 16th frames are chosen to represent the \textit{talking} action.

With these characteristics, our model may generate more accurate descriptions than ground-truths. For example, in Figure~\ref{fig:MoreResult}(b), our model explicitly indicates there is a \textit{woman}, while the ground-truth only use \textit{someone} to refer to it. In Figure~\ref{fig:MoreResult}(f), the generated sentence correctly describes the content of the video, while the ground-truth just tells it is a movie clip.

\subsection{Online video captioning}
\begin{figure}
  \centering
  \includegraphics[width=\linewidth]{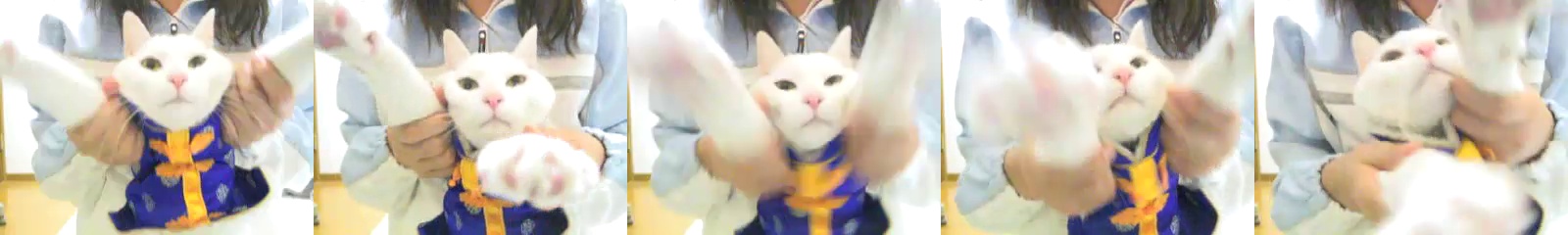}
  (a)
  {\footnotesize \textbf{a cat is licking its lips $\rightarrow$ a woman is a baby $\rightarrow$ a woman is a baby $\rightarrow$ a woman is feeding a baby $\rightarrow$ {\color{blue} a woman is playing with a kitten}}}
  \vspace{0.2em}

  \includegraphics[width=\linewidth]{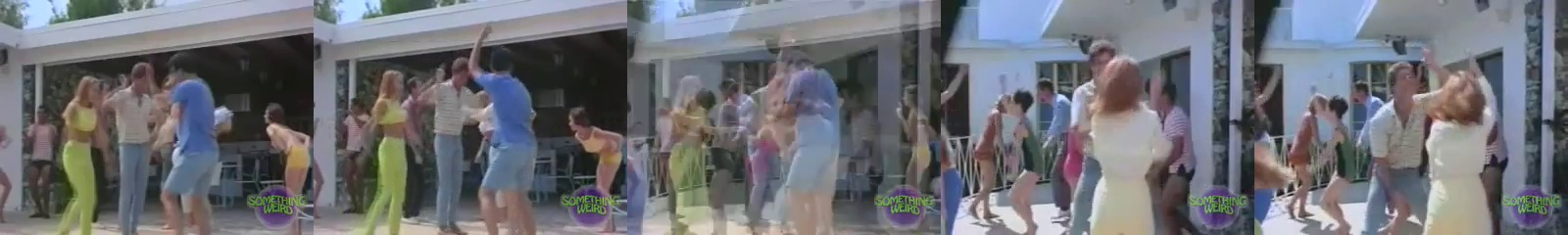}
  (b)
  {\footnotesize \textbf{a boy is running $\rightarrow$ a boy is running $\rightarrow$ a boy is running $\rightarrow$ the boys are dancing $\rightarrow$ {\color{blue} three persons are dancing}}}
  \vspace{0.2em}

  \includegraphics[width=\linewidth]{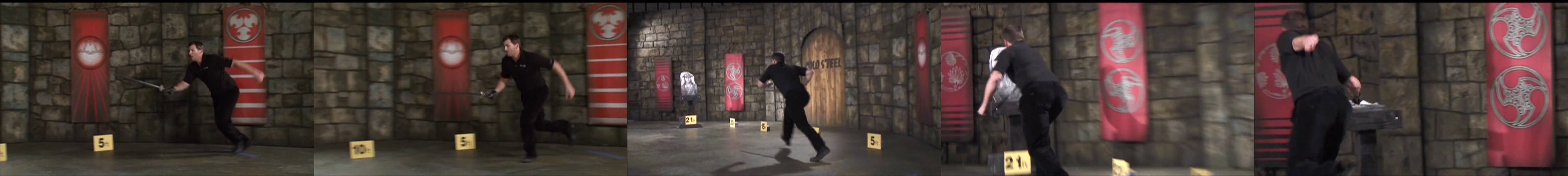}
  (c)
  {\footnotesize \textbf{a man is a sword $\rightarrow$ a boy is doing a $\rightarrow$ a man with a sword stabs a target $\rightarrow$ {\color{blue} a man is stabbing a silhouette with a sword} $\times 2$}}
\caption{Example results of online video captioning.}
\label{fig:MoreOnline}
\end{figure}

For online video captioning, we first sample frames at 1fps, and then sequentially feed the sampled frames to \textsf{PickNet}. If a certain frame is picked, the pre-trained CNN will be used to extract visual features of this frame. After that, the encoder will receive this feature, and produce a new encoded representation of the video stream up to current time. Finally, the decoder will generate a description based on the encoded representation. Figure~\ref{fig:MoreOnline} demonstrates some examples of online video captioning with the picked frames and the corresponding descriptions.
As we discussed before, the descriptions will be more appropriate and determined as the informative frames are picked. In example~(a), the most salient object in the first picked frame is the \textit{cat}, so the generated description is just about the cat. After observing enough frames, the model knows this video is about \textit{a woman is playing with a kitten}, and produces the correct description. In example~(b), the model first generates a description that \textit{a boy is running}, since the man in a blue shirt is more prominent and the motion pattern seems like \textit{running}. Along with picking the following frames, the other two persons are noticed and their actions are recognized as \textit{dancing}, hence the model produces a more accurate description that \textit{three persons are dancing}. And at the beginning of example~(c), the model is only aware of that there is a \textit{man} with a \textit{sword}, and do not know what the man is doing. After picking the third frame, it is clear that the man is \textit{stabbing} a target, then the word \textit{target} is substituted by a more precise word \textit{silhouette} when more frames are picked. The video version of online captioning results can be seen in the uploaded videos.

\end{document}